\pdfoutput=1

\documentclass[11pt]{article}

\usepackage{acl}

\usepackage{times}
\usepackage{latexsym}

\usepackage[T1]{fontenc}

\usepackage[utf8]{inputenc}

\usepackage{microtype}

\usepackage{times}
\usepackage{latexsym}
\usepackage{amsmath}
\usepackage{multirow}
\usepackage{adjustbox}
\usepackage{xcolor}
\usepackage{bm}
\usepackage{makecell}
\usepackage{fontawesome5} 
\usepackage{booktabs}

\newcommand*\mean[1]{\bar{#1}}

\definecolor{output}{HTML}{652CB3}
\definecolor{factual}{HTML}{F24726}
\definecolor{noise}{HTML}{DA0063}
\definecolor{modality}{HTML}{0CA789}
\definecolor{input}{HTML}{414BB2}
\definecolor{formatting}{HTML}{2D9BF0}

\title{Instructions for *ACL Proceedings}

\author{First Author \\
  Affiliation / Address line 1 \\
  Affiliation / Address line 2 \\
  Affiliation / Address line 3 \\
  \texttt{email@domain} \\\And
  Second Author \\
  Affiliation / Address line 1 \\
  Affiliation / Address line 2 \\
  Affiliation / Address line 3 \\
  \texttt{email@domain} \\}

\title{\textsc{Donkii}: Characterizing and Detecting Errors in Instruction-Tuning Datasets}

\author{Leon Weber-Genzel\textsuperscript{\faMountain\faRobot} \and Robert Litschko\textsuperscript{\faMountain\faRobot} \and Ekaterina Artemova\textsuperscript{\faMountain}\thanks{\ \ Now at Toloka.AI} \\ \and
         \textbf{Barbara Plank}\textsuperscript{\faMountain\kern1pt\faRobot\kern1.5pt\faCompass}\\
  \textsuperscript{\faMountain} MaiNLP, Center for Information and Language Processing, LMU Munich, Germany \\
  \textsuperscript{\faRobot} Munich Center for Machine Learning (MCML), Munich, Germany \\
  \textsuperscript{\faCompass} Department of Computer Science, IT University of Copenhagen, Denmark  \\
{\tt \{leonweber, robert.litschko, b.plank}\}@lmu.de}

\begin{document}
\maketitle
\begin{abstract}
Instruction tuning has become an integral part of training pipelines for Large Language Models (LLMs) and has been shown to yield strong performance gains.  
In an orthogonal line of research, Annotation Error Detection (AED) has emerged as a tool for detecting quality problems in gold standard labels. So far, however, the application of AED methods has been limited to classification tasks. It is an open question how well AED methods generalize to language generation settings, which are becoming more widespread via LLMs. 
In this paper, we present a first and novel benchmark for AED on instruction tuning data: \textsc{Donkii}.
It comprises three instruction-tuning datasets enriched with error annotations by experts and semi-automatic methods. 
We also provide a novel taxonomy of error types for instruction-tuning data.
We find that all three datasets contain clear errors, which sometimes propagate directly into instruction-tuned LLMs. We propose four AED baselines for the generative setting and evaluate them extensively on the newly introduced dataset. 
Our results show that the choice of the right AED method and model size is indeed crucial and derive practical recommendations for how to use AED methods to clean instruction-tuning data. 
\end{abstract}

\section{Introduction}
Recent successes in instruction tuning (InstT) have shown that Large Language Models (LLMs) can generalize to a wide range of tasks in the zero-shot setting~\citep{weiFinetunedLanguageModels2022,sanhMultitaskPromptedTraining2022,ouyangTrainingLanguageModels2022}.
InstT achieves this by training an LLM on \textit{instruction}-\textit{output} pairs, where the instruction describes the task and the output contains the expected solution to the task.
After fine-tuning on the InstT dataset and an optional reinforcement learning phase~\citep{ouyangTrainingLanguageModels2022}, LLMs are able to generalize to instructions not seen during fine-tuning.
In an orthogonal line of inquiry, researchers have studied Annotation Error Detection (AED), which allows to detect erroneous annotations in labelled datasets.
These low quality instances are then corrected or removed in a semi-automated process~\citep{vlachosActiveAnnotation2006,klieAnnotationErrorDetection2022,weber-plank-2023-activeaed}.
However, how to best apply AED for natural language generation has so far not been studied.
In this work, we combine both strands of research and ask whether AED methods can help to detect errors in InstT datasets and thus help to improve model quality by improving data quality.

\begin{figure*}[htpb]
    \centering
    \includegraphics[width=0.85\linewidth]{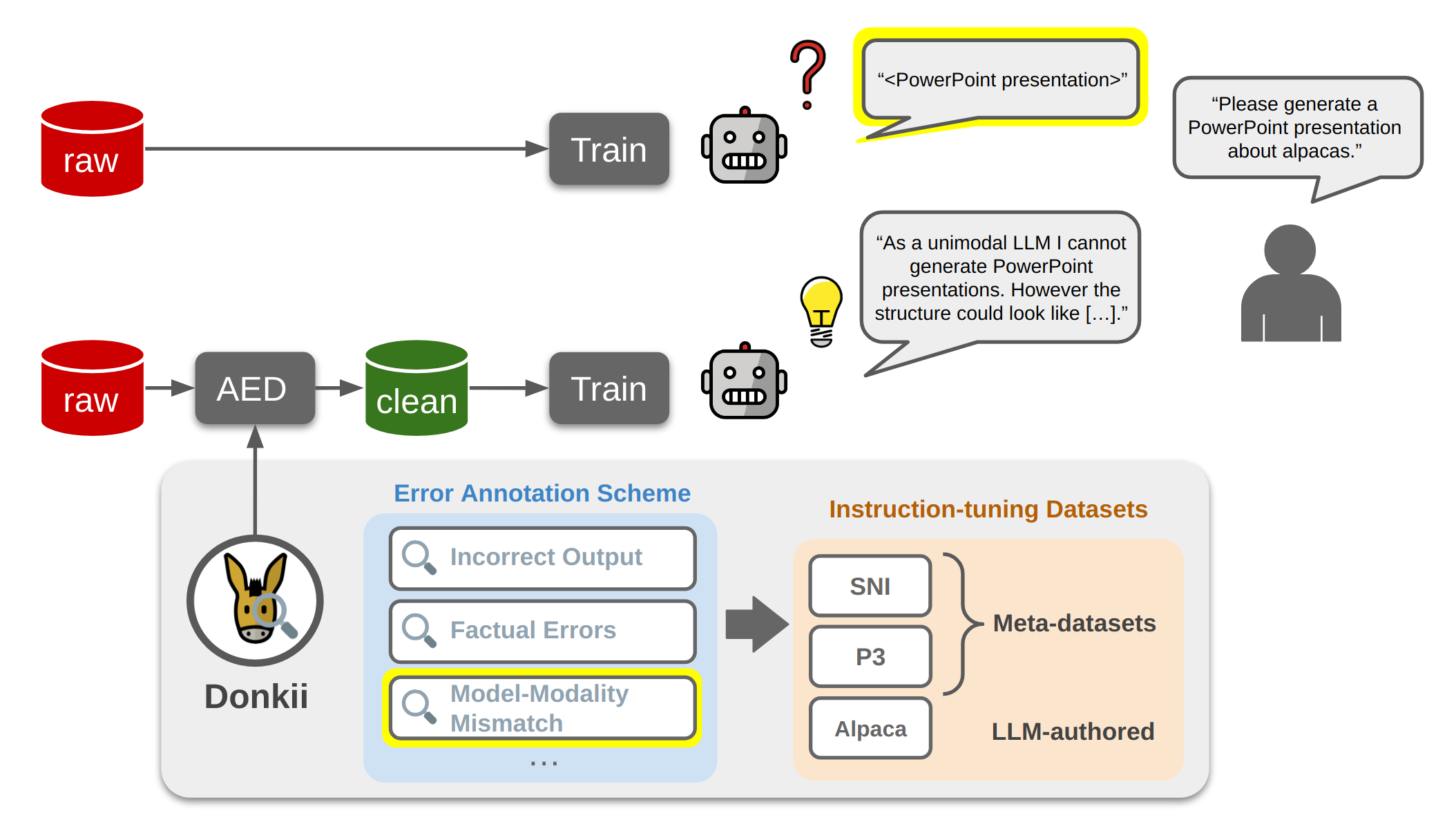}
    \caption{The Donkii dataset helps to design AED methods that can clean InstT datasets.}
    \label{fig:Donkii}
\end{figure*}

Applying AED methods to InstT datasets presents a number of challenges.
(1) The systematic development and comparison of AED methods requires datasets with annotations indicating which instances contain annotation errors.
Such datatsets are not yet available for InstT.
(2) To our knowledge, researchers have applied AED methods only in the discriminative setting~\citep{klieAnnotationErrorDetection2022} and it is not immediately clear how existing methods can be adapted to generative problems.
(3) It is not obvious what even constitutes an error in InstT.
In this work, we address these three challenges; see also Figure~\ref{fig:Donkii} for an illustration of our contributions:\footnote{Data and code are available at \url{https://github.com/mainlp/donkii}.}
(1) We present Donkii, the first instruction tuning benchmark to enable the evaluation of AED methods. Donkii contains error annotations on top of three existing InstT datasets derived from manual error annotation efforts. We also introduce a hierarchy of error types for InstT datasets; see Figure~\ref{fig:errors}.
(2) We derive four AED baselines for generative problems based on recent work on training dynamics for AED~\citep{swayamdiptaDatasetCartographyMapping2020,pleissIdentifyingMislabeledData2020}.
(3) We use Donkii to evaluate the proposed AED baselines and study the effects of model size, different types of errors, and different types of InstT data.
The results show that there is a clear best-performing AED method for InstT data among the four evaluated.

\section{Background}

\subsection{Instruction Tuning}
Instruction tuning (InstT) is an emerging paradigm that leverages natural language instructions to fine-tune language models, thereby improving zero-shot performance on unseen tasks~\citep[\textit{inter alia}]{sanhMultitaskPromptedTraining2022,ouyangTrainingLanguageModels2022,weiFinetunedLanguageModels2022,wang-etal-2022-super}.
In InstT, an LLM is fine-tuned to produce a desired output given an instruction text.
In some datasets, the instruction is further divided into a definition or prompt component, which defines the task and an optional input component~\citep{wang-etal-2022-super,taori-etal-2023-alpaca}.
In this work, we distinguish \textit{three types of InstT datasets} based on their provenance: meta-datasets, human-authored datasets and LLM-authored datasets.
The first InstT datasets were \textbf{meta-datasets}, which convert existing NLP datasets into InstT data with human-authored prompt templates~\citep{khashabi-etal-2020-unifiedqa,ye-etal-2021-crossfit,mishra-etal-2022-cross,sanhMultitaskPromptedTraining2022,weiFinetunedLanguageModels2022}.
Researchers typically construct them by writing one or more prompt templates for an existing NLP dataset.
This template is then used to transform each instance of the existing dataset into an InstT instance.
Here, we call the combination of an existing dataset and a prompt template a task.
For \textbf{human-authored} InstT datasets on the other hand, the dataset creators ask human annotators to author InstT instances~\citep{ouyangTrainingLanguageModels2022} or mine InstT instances from existing human-authored resources such as forums and wikis~\citep{zhou-et-al-2023-less}.
\textbf{LLM-authored} datasets instead are generated by LLMs.
This is typically achieved by prompting the LLM with a few examples of what InstT instances look like and instructing the model to generate new ones~\citep{wangSelfInstructAligningLanguage2022,honovich-etal-2023-unnatural,taori-etal-2023-alpaca} or by providing elaborate rules about what properties InstT instances should have~\citep{bai-etal-2022-constitutional,sun-etal-2023-principle-driven}.
Finally, dataset creators have proposed mixtures of these approaches, e.g.\ by manually correcting LLM-authored instances~\citep{ruebsamen-etal-2023-adc} or by combining instances generated by different approaches~\citep{zhou-et-al-2023-less}.
Some of these works highlight the importance of high quality data~\citep{zhou-et-al-2023-less,ruebsamen-etal-2023-adc}, but to the best of our knowledge, our study is the first to systematically evaluate AED techniques.

\subsection{Annotation Error Detection}
AED for Natural Language Processing (NLP) datasets has a long tradition, which has recently been comprehensively reviewed \citet{klieAnnotationErrorDetection2022}.
Existing AED methods can be divided into six different categories~\citep{klieAnnotationErrorDetection2022}: \textit{variation-based} methods exploit the observation that instances with similar surface forms tend to have the same label~\citep{dickinsonDetectingErrorsPartofSpeech2003,larsonInconsistenciesCrowdsourcedSlotFilling2020a}.
\textit{Model-based} methods use a cross-validation scheme to generate predictions for the whole dataset and then use these predictions to flag errors, e.g.\ by highlighting instances where the predicted label is different from the one assigned ~\citep{amiriSpottingSpuriousData2018,yaghoub-zadeh-fardStudyIncorrectParaphrases2019}.
\textit{Training-dynamics-based} approaches compute statistics on quantities collected during training~\citep{swayamdiptaDatasetCartographyMapping2020,pleissIdentifyingMislabeledData2020,siddiquiMetadataArchaeologyUnearthing2022}.
\textit{Vector-space-proximity-based} methods assume that instances that are close in a suitable vector space should have the same label~\citep{larsonOutlierDetectionImproved2019,grivasNotCuteStroke2020}.
\textit{Ensemble-based} methods use statistics of the predictions of ensemble members to find errors~\citep{altTACREDRevisitedThorough2020,varshneyILDAEInstanceLevelDifficulty2022} and rule-based approaches rely on manually defined rules to spot erroneous instances~\citep{kvetonSemiAutomaticDetection2002}.
In this work, we focus on training dynamics because they performed well in a recent evaluation~\citep{klieAnnotationErrorDetection2022}, can be relatively easily adapted to generative settings and have a low computational overhead. We leave the evaluation of other types of methods to future work.
An orthogonal classification of AED methods is into flaggers and scorers~\citep{klieAnnotationErrorDetection2022}.
Flaggers model AED as a binary classification task (error vs non-error) and scorers assign an error score to each instance that reflects the likelihood that the instance contains an annotation error.
In this work, we focus on scorers, because the ranking induced by them allows more fine-grained decisions about which instances to inspect~\citep{weber-plank-2023-activeaed} and they can be converted to flaggers by choosing an appropriate threshold~\citep{swayamdiptaDatasetCartographyMapping2020}.

\section{Proposed AED baselines for text generation datasets} \label{sec:methods}
We present four AED baselines for text generation datasets.
For this, we adapt methods based on training dynamics that were previously used for AED in classification problems~\citep{swayamdiptaDatasetCartographyMapping2020,pleissIdentifyingMislabeledData2020}.
We chose these methods because they performed well in earlier work~\citep{klieAnnotationErrorDetection2022,weber-plank-2023-activeaed} and because their adaptation to generative settings is relatively straight-forward.
All four methods assign an error score to an instance, with a higher score ideally reflecting a higher probability of an incorrect annotation.
All scores use the probabilities $p_{e,l}$ that the model assigned to the token $l$ of the instance's output sequence at epoch $e$ during training.
We propose the following measures:
(1) \textbf{Perplexity}, which is the epoch-averaged perplexity of the instance based on $p_{e,l}$:
\begin{equation}
    \text{PPL} = \frac{1}{E} \sum_{e=1}^E \text{ppl}_e,
\end{equation}
where $E$ is the number of epochs and $\text{ppl}_e$ the perplexity at epoch $e$.
(2) The (negative) \textbf{average probability}, determined by averaging $p_{e,l}$:
\begin{equation}
    P_\mu = - \frac{1}{E} \sum_{e=1}^E \frac{1}{L} \sum_{l=1}^{L} p_{e,l},
\end{equation}
where $L$ is the number of tokens in the output sequence.

(3) The (negative) \textbf{minimum probability}, derived from the minimum of $p_{e,l}$:
\begin{equation}
    P_\text{min} = - \frac{1}{E} \sum_{e=1}^E \min_{l=1}^{L} p_{e,l}.
\end{equation}
(4) The (negative) \textbf{Area-under-the-Margin score} (AUM)~\citep{pleissIdentifyingMislabeledData2020}, which we adapt to the generative setting by calculating it for each token in the output sequence and averaging the resulting scores:
\begin{equation}
    \text{AUM} = \frac{1}{E} \sum_{e=1}^{E} \frac{1}{L} \sum_{l=1}^{L} \max_{y_l' \neq y_l} p_{e}(y_l'| x_l) - p_{e,l},
\end{equation}
where $y_l$ is the token at position $l$ and  $\max_{y_l' \neq y_l} p_{e}(y_l'| x_l)$ is the maximum probability assigned by the model at epoch $e$ for position $l$ excluding the assigned token. 
In addition, we consider a variant of each score that uses only the last epoch; see \S\ref{ssec:results} for the results of this ablation.
\begin{figure*}[htbp]
    \centering
    \includegraphics[width=1.0\linewidth]{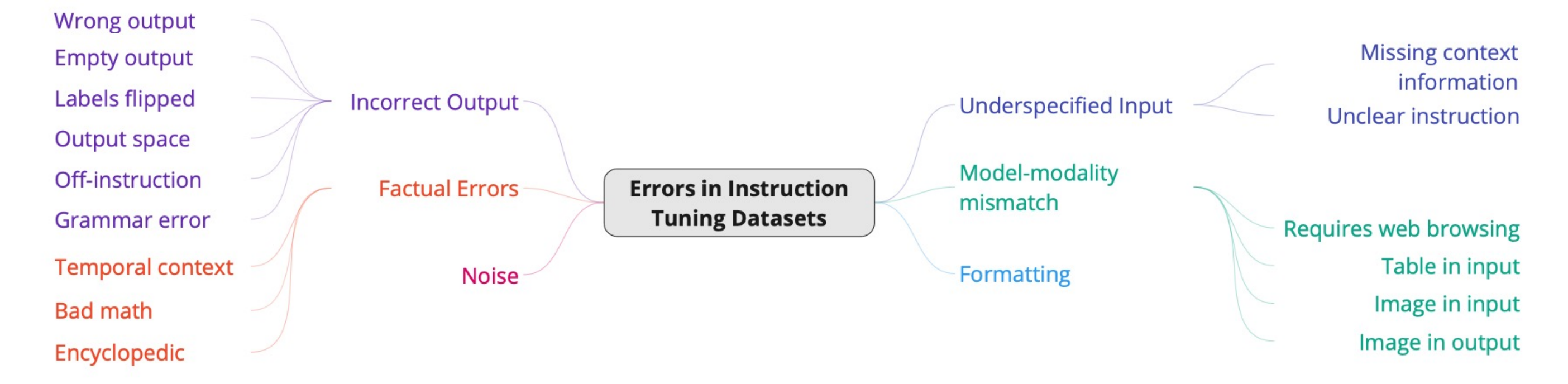}
    \caption{The \textsc{Donkii} taxonomy of six error categories, four of which are further divided into more specific subcategories.}
    \label{fig:errors}
\end{figure*}

\section{Datasets and Error Types} \label{sec:datasets}
We describe Donkii's three different data sources that we have enriched with annotations of erroneous instances: P3-Donkii, SNI-Donkii, and ADC-Donkii.\footnote{See Appendix~\ref{sec:data_statement} for our data statement.}
Each is based on an existing InstT dataset and on manual inspection of the errors in that dataset: P3-Donkii is derived from the meta-dataset Public Pool of Prompts~\citep{sanhMultitaskPromptedTraining2022}, SNI-Donkii from the meta-dataset Super-Natural Instructions~\citep{wang-etal-2022-super}, and ADC-Donkii from the LLM-authored dataset Alpaca~\citep{taori-etal-2023-alpaca} and its partially corrected version AlpacaDatasetCleaned~\citep{ruebsamen-etal-2023-adc}.
We enrich each of these datasets with labels indicating which instances contain errors, using different mixtures of expert annotation and programmatic analysis of the source data.
For each dataset, we construct three different sets of instances: $\mathcal{X}^*$ which contains no errors, $\mathcal{X}_\text{err}$, for which we know that it contains errors, and $\mathcal{X}_\text{unk}$, for which we do not know if it contains errors.
We evaluate AED methods by their ability of discriminating $\mathcal{X}^*$ from $\mathcal{X}_\text{err}$.
We exclude $\mathcal{X}_\text{unk}$ from evaluation, as we do not exhaustively annotate the datasets with errors due to their sheer size and resource availability; see \S\ref{ssec:evaluation} for details.

\subsection{P3-Donkii}
Public Pool of Prompts (P3)~\citep{sanhMultitaskPromptedTraining2022} is a meta-dataset for InstT which was created by asking researchers and open-source contributors to transform existing datasets by writing prompts using the InstT templating engine \textit{promptsource}~\citep{bach-etal-2022-promptsource}.
We construct the P3-Donkii dataset by introducing different types of synthetic errors into the P3 data.
We use this synthetic setup for P3 so that we have full control over the the number and types of errors in the dataset.\footnote{An alternative approach would have been a full manual annotation of P3 which was out of reach because of the large size of the dataset.}
To find realistic error classes, we first detect existing errors in the P3 data.
We use PPL to assign error scores to tasks in P3, employing both mean and median aggregation (see \S\ref{ssec:results} for details on this).\footnote{In principle, this semi-automatic process of finding error categories with PPL could bias our evaluation results. However, our purely manual analysis of SNI shows similar error categories, so we are confident that the introduced bias is minimal.}
We then manually inspect the top 20 highest scoring tasks by looking at their highest scoring instances. 
In our manual inspection, we found the following types of problems:

\noindent \textbf{Empty output}: The output is an empty string where it should not be. 

\noindent \textbf{Incorrect output}: The output contains severe orthographic or factual errors.

\noindent \textbf{Missing context information}: The prompt is truncated during preprocessing. This can make crucial information unavailable, e.g. missing context in extractive QA.

We then correct the errors that we found in the 20 tasks in P3.
We rebuild the empty output data from scratch using \textit{promptsource} and verify that the output strings are not empty.
We remove tasks that contain a high number of low quality outputs.
We discard instances that do not fit within the set maximum length, so that there is no missing information.
Finally, we reintroduce the detected errors in a controlled manner by modelling them synthetically. 
For each type of error, we randomly sample five tasks and perturb their instances:

\noindent \textbf{Empty output}: We replace the output in all instances for the task with an empty string.

\noindent \textbf{Low quality output}: For each instance of the task, with a probability of $0.5$, we replace the gold-standard output with output generated by prompting Llama-7b~\citep{touvron-etal-2023-llama}. 

\noindent \textbf{Missing information}: For each instance of the task, with a probability of $0.5$, we truncate the gold-standard prompt to half of its original length.

\noindent \textbf{Flipped output}: For each instance, with a probability of $0.5$, we replace the output with the output of another instance. This is a widely used perturbation used in AED research~\citep{klieAnnotationErrorDetection2022}, which we adapted to InstT datasets.

We collect all perturbed instances in $\mathcal{X}_\text{error}$ (the set of instances the AED methods should detect) and those of the same original\footnote{Note, that for all tasks during our initial exploration of the dataset, we use the corrected error-free version of the task.} unperturbed task in $\mathcal{X}^*$ (the set of instances which should not be detected by the AED methods).
The instances of unperturbed tasks constitute $\mathcal{X}_\text{unk}$.

\begin{table*}[htbp]
      \centering
      \adjustbox{max width=\linewidth}{%
        \begin{tabular}{lcccccccccc}
              \toprule
              & \multicolumn{1}{l}{Source data} & \multicolumn{1}{l}{$|\mathcal{X}_\text{unk}|$} &   \multicolumn{1}{l}{$|\mathcal{X}^*|$} & \multicolumn{1}{l}{$|\mathcal{X}_\text{err}|$}  & \multicolumn{1}{l}{$|\mathcal{T}|$} &  \multicolumn{1}{l}{$|\mathcal{T}_\text{err}|$} & \multicolumn{1}{l}{$\mean{L}_\text{inp}$} & \multicolumn{1}{l}{$\mean{L}_\text{out}$} & \multicolumn{1}{l}{Err} & \multicolumn{1}{l}{Prov} \\
              \midrule
        P3    & \citet{sanhMultitaskPromptedTraining2022}  & 399,472 & 12,237 & 12,237 &   417   & 20 & 118 & 9 & Syn. & Meta  \\
        \midrule
        SNI   & \citet{wang-etal-2022-super}  &  101,783  &  1,088     &   585    &   1,613          &  17 & 165 & 6 & Nat. & Meta  \\
        \midrule
        ADC   &  \shortstack{\citet{taori-etal-2023-alpaca} \\ \citep{ruebsamen-etal-2023-adc}} & 48,425 & 173 & 146  &     - & -  & 15 & 44 & Nat & LLM  \\
        \bottomrule
        \end{tabular}%
        }
      \caption{Statistics for the three Donkii datasets. $|\mathcal{T}|$ denotes the total number of tasks, and $|\mathcal{T}_\text{err}|$ the number of tasks with at least one instance with an error. Note, that ADC does not provide a grouping of instances into tasks. $\mean{L}_\text{inp}$/$\mean{L}_\text{out}$ denotes the average input/output length in white-space-delimited tokens. `Err' is the type of error (synthetic or naturally ocurring) and `Prov' the provenance (meta-dataset vs LLM-authored). `Lic' is the license under which the authors published their data. }
      \label{tab:datasets}%
    \end{table*}%

\subsection{SNI-Donkii}
\label{ssec:sni-aed}
We construct SNI-Donkii by mining errors that arose during the creation of the Super-natural instructions (SNI)~\citep{wang-etal-2022-super} dataset and have been corrected for the current version of SNI.
SNI\footnote{\url{https://github.com/allenai/natural-instructions}} is a meta-dataset for InstT, created by a large number of researchers by transforming existing datasets into InstT tasks.
It contains a total of 1,616 tasks covering a wide range of NLP tasks such as  question answering, text classification, sentiment analysis, textual entailment, and summarization.
The project implemented quality control through peer review conducted via GitHub\footnote{\url{https://github.com/allenai/natural-instructions/commits/master}} and a crowd-sourced evaluation.
We create SNI-Donkii by comparing the version of each task before the final round of peer review (the first version uploaded to GitHub) and after peer review (the version on GitHub at the time of writing). 
From the 1,613 tasks that we were able to download without error, we collect all 455 tasks $t \in \mathcal{T}$ where the output of at least one instance changed.
For 17 of these tasks, all expert annotators (two co-authors and one NLP MSc student) agreed\footnote{We opted for a roundtable discussion rather than majority voting because we found that annotating errors in InstT datasets is a difficult task. Even though we relied exclusively on expert annotators, they sometimes missed crucial details about instances and revised their annotations during the discussion. A disadvantage of this discussion-based setup is that we cannot reliably estimate inter-annotator agreement.} that at least 90\% of the changed instances contain an error.
See Table~\ref{tab:pairwise_example} for an example of the annotation task and Table~\ref{tab:examples} for examples of found errors.
For the annotation guidelines see Appendix~\ref{sec:guidelines_sni}.
From this annotation, we construct SNI-Donkii as follows:
For all tasks without errors, we add 64 instances of the latest version -- or less if the task contains fewer than 64 instances -- of it to $\mathcal{X}_\text{unk}$.
For the erroneous tasks, we add 64 changed instances from the oldest version to $\mathcal{X}_\text{err}$ and 64 from the newest version to $\mathcal{X}^*$.\footnote{We chose $64$ instances because \citet{wang-etal-2022-super} find that performance plateaus with more instances per task.}
When we had fewer than 64 updated instances, we put them all into $\mathcal{X}_\text{err}$.
In this case, we filled up $\mathcal{X}_\text{unk}$ with extra instances from the oldest version to keep the number of instances for each task about the same.
Table~\ref{tab:datasets} gives the statistics of the resulting dataset.

\begin{table}[htbp]
\adjustbox{{max width=0.99\linewidth}}{%
\centering
\begin{tabular}{p{8cm}}
\toprule
\textbf{Instruction 1}: Name two deserts in the Sahara.	\\
\textbf{Input 1}: \\
\textbf{Output 1}: The two deserts in the Sahara are the Great Western Erg and the Great Eastern Erg.	 \\ \midrule

\textbf{Instruction 2}: Recognize the following bird's species.	\\
\textbf{Input 2}: \textcolor{modality}{<Image of a bird>}	\\
\textbf{Output 2}: Western Great Egret (Ardea alba maxima). \\ \midrule
\textbf{Label}: 1 is better than 2	\\		
\textbf{Error category}: \textcolor{modality}{Image in input} \\
\bottomrule
\end{tabular}%
}
\caption{Example for the pair-wise annotation task that we used to flag errors in SNI-Donkii and ADC-Donkii.}
\label{tab:pairwise_example}%
\end{table}%

\begin{table*}[htbp]
\centering
\adjustbox{{max width=0.99\linewidth}}{%
\tiny
\begin{tabular}{p{1cm}lp{2.5cm}p{3cm}p{6cm}}
\toprule
Error & Source & Instruction & Input & Output (shortened) \\
\midrule
\textcolor{modality}{Requires web browsing} & Alpaca &  Summarize the New York Times article about the rise of AI. & \textcolor{modality}{\texttt{www.nytimes.com/2020/06/16/ opinion/ai-growth.html}} & The New York Times article titled ``The Rise of Artificial Intelligence'' looks at the growing impact of AI on the world today. <...> \\
\midrule
\textcolor{factual}{Bad math} & Alpaca &  Calculate the mean of the numbers 3, 4, 6, 10, 11 & & The mean of the numbers 3, 4, 6, 10, 11 is 6.4. This is calculated by adding the numbers together and then dividing by the number of values (5). (3 + 4 + 6 + 10 + 11 = 34; 34 / 5 = \textcolor{factual}{6.4}). \\
\midrule
\makecell[l]{\textcolor{input}{Unclear} \\ \textcolor{input}{instruction}} &  Alpaca   &  \textcolor{input}{Sort the following 10 fruits into three categories.}  & Bananas, Apples, Pineapples, Plums, Cherries, Peaches, Grapes, Watermelons, Mangoes, Oranges  & Category 1: Bananas, Apples, Pineapples, Plums, Cherries;\\
\midrule
\makecell[l]{\textcolor{formatting}{Formatting}} &  SNI   &  [...] & story: I went down to the tidepool to watch the tide roll out. [...]
 selected sentence: I went down to the tidepool to watch the tide roll out. &   I decide \textcolor{formatting}{togotothe} tidepool >Causes/Enables> I \textcolor{formatting}{gotothe} tidepool  \\
 \midrule
\textcolor{output}{Labels flipped} & SNI    &  You are given two sentences (Sentence1 and Sentence2). Answer ``Yes'' if these sentences are a paraphrase of one another, otherwise answer ``No''. & Sentence1: The broader Standard \& Poor 's 500 Index .SP> gained 3 points , or 0.39 percent , at 924 ., 

Sentence2: The technology-laced Nasdaq Composite Index .IXIC rose 6 points , or 0.41 percent , to 1,498 . & \textcolor{output}{Yes}  \\
\bottomrule
\end{tabular}%
}
\caption{Examples of some error categories of the Donkii taxonomy. }
\label{tab:examples}%
\end{table*}%

\subsection{ADC-Donkii}
\label{ssec:alpaca-aed}
Alpaca~\citep{taori-etal-2023-alpaca} is an LLM-authored dataset constructed by following the self-instruct recipe proposed by~\citet{wangSelfInstructAligningLanguage2022}.
The creators of Alpaca repeatedly prompted text-davinci-003\footnote{\url{https://platform.openai.com/docs/models}} with in-context examples of InstT instances sampled from a pool of human and LLM-authored instances and asked the LLM to provide a new instance.
This yielded a dataset of 52,000 different InstT instances. 
In a separate effort called AlpacaDataCleaned (ADC)~\citep{ruebsamen-etal-2023-adc}, members of the open source community corrected errors in the Alpaca data using a mixture of manual and rule-based annotation.\footnote{\url{https://github.com/gururise/AlpacaDataCleaned/issues/31}}
To construct ADC-Donkii, we collect 300 instances from Alpaca that do not occur in ADC and pair each of them with the instance with the closest BM25 score from ADC.
Using these pairs, three of this study's authors manually annotate whether one of the two instances is clearly preferable because the other contains at least one error.
The annotation guidelines can be found in Appendix~\ref{sec:guidelines_adc}.
As with SNI, we do not disclose which instances are from Alpaca and which are from ADC to avoid introducing unnecessary bias.
If, after a roundtable discussion, all three annotators agree that one of the two instances is preferable, we add it to $\mathcal{X}^*$ and the other to $\mathcal{X}_\text{err}$.
We add all other instances from Alpaca and ADC to $\mathcal{X}_\text{unk}$.
Table~\ref{tab:datasets} shows the statistics for the resulting dataset.

\subsection{Error categories}
During annotation, we identified six main categories of errors, each with several subcategories. 
Note, that these error categories are not exhaustive and are observed in the annotated sample, rather than encompassing all possible categories of errors.
A more detailed and nuanced analysis of possible errors in instruction tuning data is the subject of future work. 
The proposed hierarchy is shown in \autoref{fig:errors}; a sample from Donkii errors is shown in \autoref{tab:examples}. 
More examples for each category can be found in the \autoref{sec:error_appendix}.
The error categories are the following:
\noindent \textbf{\textcolor{output}{Incorrect output}}: Problems are observed in the output.
This may include providing inaccurate or incorrect output, such as providing a three-letter abbreviation when a two-letter abbreviation was requested.
Other problems in this category include not providing any output at all, reversing the label in binary classification tasks, and providing output that is in the wrong output space, such as answering a/b/c/d in a multiple choice question when the options are listed as 1/2/3/4.
In addition, the output may be an off-instruction response that is related to the instruction but does not follow it, for example, responding with a code example that can calculate an average instead of directly outputting the average of the given numbers as requested. 
Finally, the output may contain ungrammatical text.
\noindent \textbf{\textcolor{factual}{Factual knowledge and mathematics}}: This category covers outputs that may be time-dependent, contain factual errors, or contain incorrect arithmetic. 
\noindent \textbf{\textcolor{noise}{Noise}}: Instances in which the instruction, input, or output contains some form of noise. 
This noise can range from \texttt{NoInput} stubs to duplicating the instruction in the output.
\noindent \textbf{\textcolor{input}{Underspecified input}}: Instances in which the instruction and input do not provide sufficient information to complete the task. 
For example, a task may ask to find the average of a set of numbers without giving the actual numbers. 
This category also includes cases where the instruction is unclear and cannot be completed correctly due to a lack of specification. 
For example, a task may require classifying data points into multiple categories without explicitly describing the semantics of the categories or providing the data points.
We argue that these instances are errors because the LLM should ask the user for the necessary input rather than assuming (i.e. `hallucinating') input.
\noindent \textbf{\textcolor{modality}{Model-modality mismatch}}: Instances that require additional modalities are placed in this category, to reflect that the examined InstT datasets are used for text-only LLMs. 
Unsupported modalities may include tables, images (as in \autoref{tab:pairwise_example}), and the use of additional tools to browse and retrieve information from the web.
\noindent \textbf{\textcolor{formatting}{Formatting}}: Instances with corrupted formatting, such as missing white spaces and the use of punctuation instead of white spaces.
In general, the identified errors are very similar to known errors made by models for open-ended text generation \cite{dou-etal-2022-gpt,ge2022tgea}, with less emphasis on issues of syntax and word choice. 

\section{How well does AED do in Instruction Tuning data?}
In this section, we provide baseline results for error detection in InstT datasets.
To do this, we evaluate the baselines introduced in \S\ref{sec:methods} using the Donkii datasets proposed in \S\ref{sec:datasets}.

\begin{table*}[htbp]
\adjustbox{{max width=\linewidth}}{%
  \centering
    \begin{tabular}{l|l|llll|llll|llll|llll}
    \toprule
    \multicolumn{1}{c}{}      & \multicolumn{1}{c}{} & \multicolumn{4}{c}{small}     & \multicolumn{4}{c}{base}      & \multicolumn{4}{c}{large}     & \multicolumn{4}{c}{xl} \\
      & rand & PPL & $P_\mu$ & $P_\textit{min}$ & AUM &  PPL & $P_\mu$ & $P_\textit{min}$ & AUM & PPL & $P_\mu$ & $P_\textit{min}$ & AUM &  PPL & $P_\mu$ & $P_\textit{min}$ & AUM \\
          \midrule
    P3   &  $50.0$ & $73.6_\textit{4.4}$ & $\bm{84.3_\textit{1.0}}$ & $51.3_\textit{0.8}$ & $52.7_\textit{0.7}$ & $76.1_\textit{0.6}$ & $81.7_\textit{0.3}$ & $51.2_\textit{0.0}$ & $53.6_\textit{0.1}$ & $70.7_\textit{3.5}$ & $77.4_\textit{0.4}$ & $49.4_\textit{0.6}$ & $52.2_\textit{0.2}$
 & $61.3_\textit{0.0}$ & $66.0_\textit{0.0}$ & $58.6_\textit{0.0}$ & $46.0_\textit{0.0}$ \\
    SNI & $34.9$ & $30.7_\textit{0.2}$  &  $30.3_\textit{0.2}$ &    $27.9_\textit{0.1}$  &  $28.7_\textit{0.1}$ &   $31.8_\textit{0.4}$  &  $42.4_\textit{0.9}$ &    $30.2_\textit{0.3}$  &  $29.8_\textit{0.1}$ &   $33.4_\textit{1.0}$  &  $\bm{48.1_\textit{1.7}}$ &    $34.5_\textit{0.9}$  &  $34.1_\textit{0.6}$ &    $33.0_\textit{0.7}$  &  $38.9_\textit{0.2}$ &   $32.2_\textit{0.2}$  & $31.6_\textit{0.4}$ \\
    ADC & $45.1$ &  $54.5_\textit{0.3}$ & $\bm{55.4_\textit{0.3}}$ & $50.3_\textit{0.5}$ & $47.9_\textit{0.7}$ & $53.7_\textit{0.21}$ & $51.9_\textit{0.5}$ & $50.8_\textit{0.7}$ & $51.4_\textit{0.7}$ & $52.2_\textit{0.4}$ & $48.9_\textit{0.2}$ & $47.5_\textit{0.5}$ & $47.2_\textit{1.2}$ & $53.1_\textit{0.0}$ & $53.1_\textit{0.0}$ & $51.3_\textit{0.0}$ & $47.3_\textit{0.0}$ \\
    \midrule
    AVG & $43.3$ & $52.9$ & $56.7$ & $43.2$ & $43.1$ & $53.9$ & $\bm{58.7}$ & $44.1$ & $44.9$ & $52.1$ & $58.1$ & $43.8$ & $44.5$ & $49.1$ & $52.7$ & $47.4$ & $41.6$ \\

    \bottomrule
    \end{tabular}%
}
  \caption{Results of four different AED methods applied to the Donkii datasets. All scores are Average Precision in percent. The larger number is the mean across three seeds and the smaller number the standard deviation. The best result per dataset is in bold. Rand is the random baseline.}
  \label{tab:results}%
\end{table*}%

\subsection{Evaluation protocol}
\label{ssec:evaluation}
We follow the evaluation protocol for scoring-based AED methods for classification tasks of \citet{klieAnnotationErrorDetection2022} and \citet{chong-etal-2022-detecting} -- with one modification.
We follow the protocol by treating the problem as a ranking task, where an AED model assigns an error score to each instance $x_i$.
However, unlike \citet{klieAnnotationErrorDetection2022} and \citet{chong-etal-2022-detecting}, we have three sets of instances instead of two:  $\mathcal{X}^*$, which contains few to no errors, $\mathcal{X}_\text{err}$ which contains many errors, and $\mathcal{X}_\text{unk}$ for which we do not know the proportion of errors.
We judge the quality of the ranking by how well it distinguishes between $\mathcal{X}^*$ and $\mathcal{X}_\text{err}$ and ignore  $\mathcal{X}_\text{unk}$ during evaluation.
Note that while we use only $\mathcal{X}^*$ and $\mathcal{X}_\text{err}$ for evaluation, we train on $x_i \in \mathcal{X}^* \cup \mathcal{X}_\text{err} \cup \mathcal{X}_\text{unk}$.
We use average precision (AP), i.e.\ the area under the precision-recall curve, implemented with scikit-learn~\citep{scikit-learn} to score the rankings and use $\frac{|\mathcal{X}_\text{err}|}{|\mathcal{X}^*| + |\mathcal{X}_\text{err}|}$  as an estimator for the random baseline~\citep{bestgen-2015-exact}.
We conduct all experiments with four models of different sizes from the T5 family\footnote{\url{https://huggingface.co/google/t5-base-lm-adapt}}~\citep{raffel-2020-exploring} in the version that \citet{lester-etal-2021-power} continually fine-tuned as language models.
We chose T5 because it has worked well in previous InstT work~\citep{sanhMultitaskPromptedTraining2022,weiFinetunedLanguageModels2022,wang-etal-2022-super}.
See Appendix~\ref{sec:hyperparameters} for the hyperparameters.
We repeat all experiments with three different seeds and report the mean and standard deviation of the results.

\subsection{Results}
\label{ssec:results}
The results can be found in Table~\ref{tab:results}.
On average, $P_\mu$ (average probability) performs the best across all model sizes, with PPL coming in second.
AUM is tied for the third place with $P_\text{min}$, each outperforming the other for two of the four model sizes.
This ranking is relatively stable for each individual dataset and the best configuration always uses $P_\mu$.
We conclude from this that $P_\mu$ clearly emerges as the best performing baseline for AED in our natural language generation setup.
This shows the striking benefits in term of simplicity and effectiveness of our proposed $P_\mu$ metric.
The improvement over the random baseline is relatively large at over 34 percentage points (pp) for P3-Donkii but more modest for SNI-Donkii and ADC-Donkii at 13.2 pp and 10.3 pp respectively.
This is probably due to the fact that synthetically introduced errors are generally easier to detect than naturally occurring ones~\citep{klieAnnotationErrorDetection2022}.
For \textbf{model size}, small is the best for P3 and ADC, while large is the best for SNI. 
On average, base and large perform best, while small also performs surprisingly well.
Therefore, for a new InstT dataset, we recommend starting with a base-sized model for efficiency reasons.
%

\begin{table}[htbp]
  \centering
    \adjustbox{{max width=\linewidth}}{%
    \begin{tabular}{llllll}
    \toprule
    \textbf{P3} &  \textcolor{output}{out} ($9777$)  & \textcolor{input}{inp} ($2460$) & - & - & - \\
    \midrule
    rand  & $50.0$ & $50.0$  & - & - & - \\
    $P_\mu$ & $89.4_\textit{0.9}$ & $68.0_\textit{0.1}$ & - & - & - \\
    \bottomrule
    \textbf{ADC}  & \textcolor{output}{out} ($13$)  & \textcolor{input}{inp} ($13$) & \textcolor{noise}{noi} ($77$) & \textcolor{factual}{fac} ($14$)  & \multicolumn{1}{l}{\textcolor{modality}{mul} (29)} \\
    \midrule
    rand  & $37.0$ & $48.0$ & $48.4$ &  $29.8$ & $50.9$ \\
    $P_\mu$ & $62.6_\textit{0.8}$ & $72.2_\textit{0.2}$ & $49.8_\textit{0.4}$ & $55.7_\textit{0.8}$ & $61.5_\textit{0.5}$ \\
    \bottomrule
    \textbf{SNI}   & \textcolor{output}{out}  &  \textcolor{formatting}{form} ($64$) & \textcolor{noise}{noi} ($2$) & - & \textcolor{modality}{mul} ($3$)\\
    \midrule
    rand  & $38.2$ & $50.0$   & $3.0$ & - & $2.3$ \\
    $P_\mu$ & $51.7_\textit{1.7} $ & $51.9_\textit{0.9}$ & $30.6_\textit{8.6}$ & & $14.9_\textit{3.9}$   \\
    \bottomrule
    \end{tabular}%
    }
  \caption{Results per error category. All scores are AP (higher is better) in percent of $P_\mu$ using the best performing model size for the dataset. The category names are abbreviated: \textcolor{output}{out: incorrect output}, \textcolor{input}{inp: underspecified input}, \textcolor{noise}{noi: noise}, \textcolor{factual}{fac: factual error}, \textcolor{modality}{mul: multi-modality}, \textcolor{formatting}{form: formatting}. The number in brackets gives the number of instances per category.}
  \label{tab:results_categories}%
\end{table}%

We analyse the performance of the different scorers per annotated \textbf{error category}. 
For each dataset, we use $P_\mu$ with the respective best performing model size.
The results can be found in Table~\ref{tab:results_categories}.
Interestingly, the results differ strongly across error categories and dataset.
$P_\mu$ outperforms the random baseline for all but two categories, which are noisy instances in ADC-Donkii and formatting errors in SNI-Donkii.
Surprisingly, other configurations, which on average perform worse than $P_\mu$, are able to beat the random baseline for these error types with the respectively best scorers outperforming random by $18.1$/$13.7$ pp for noise/formatting.

\noindent \textbf{On instance vs task-level and epoch aggregation}
Our annotation of SNI and P3 showed that errors in meta-datasets often affect a large proportion of all instances for a given task.\footnote{This observation motivated our annotation efforts for P3 and SNI.}
We wondered whether we could exploit this property by \textbf{aggregating error scores across all instances for a given task} and thus perform AED on tasks rather than instances.
For this, we conducted additional experiments using SNI-Donkii, where we computed two scores for each task by taking the mean and median across all instances for the given task.
We then follow the same ranking-based evaluation protocol as for individual instances.
Here, we observe a slightly different ordering of methods, with PPL achieving the highest score.
On average, the aggregation by median yielded higher scores than aggregation by mean.
The absolute AP is much higher than for single instance error detection at $69.3\%$ (vs $48.1\%$), suggesting that task aggregation may be useful for detecting systematic errors in meta-datasets.
We also examine the effect of \textbf{aggregating scores over all epochs}.
For this, we ablate the epoch aggregation by using the final logits directly to compute the AED scores.
For each dataset, we compute the difference between the best performing size-score combination with and without epoch aggregation.
We find that the scores drop by 1.3/3.9/1.2 percentage points AP for P3/SNI/ADC respectively \textit{without} aggregation over epochs.
This further supports the observation that averaging AED scores over epochs generally improves performance~\citep{swayamdiptaDatasetCartographyMapping2020,pleissIdentifyingMislabeledData2020,weber-plank-2023-activeaed}.

\section{Conclusion}
This work presents the first study on annotation error detection for generation tasks, in particular, instruction tuning data.
Despite the popularity of InsT, there are no evaluation datasets for AED with marked errors.
Therefore, we present Donkii, a suite of three existing InstT datasets enriched with novel error annotations and an error taxonomy derived from manual annotation efforts.
We propose four different AED methods for generative models and systematically evaluate them on the Donkii datasets.
We find that there is a clear best performing method for single instances with $P_\mu$ and for task-level AED with PPL.
In any case, the choice of model size is critical for optimal AED performance.
In Appendix~\ref{sec:impact} we report on preliminary experiments in which we investigated how annotation errors impact downstream performance.
For future work, we plan to apply AED methods to more structured generative meta-datasets such as \citet{huguet-cabot-navigli-2021-rebel-relation} or \citet{friesBigBioFrameworkDataCentric2022}.

\section*{Limitations}

\paragraph{Identified errors in InsT datasets.} We acknowledge that the error categories we have identified are not exhaustive. This is because the current errors have been annotated based on manual examination of medium-sized samples. We also acknowledge that our error category does not cover issues related to toxicity, hallucinations, and safety, as we believe that these issues are so important that they require specialized treatment in more focused work~\citep[\textit{inter alia}]{gehman-etal-2020-realtoxicityprompts,sap-etal-2022-annotators,raunak-etal-2021-curious,dziri-etal-2022-origin,greshakeNotWhatYou2023}.
\paragraph{Small sample size for individual categories}
We invested significant manual effort in annotation, but strongly favoured precision over quantity, with three expert annotators first labeling each sample individually and then discussing the results.
As a result, the number of errors found per category is moderate to small (see Table~\ref{tab:results_categories}).
We believe that an even larger annotation effort would be required in the future to ensure that all findings on error categories are robust.

\section*{Ethics \& Broader Impact}
Instruction-tuned LLMs have been widely adopted by non-expert users~\citep{openai-2023-gpt4}.
We believe that this makes fine-grained control over the model outputs and thus, by extension, over the content of the InstT dataset an ethical imperative.
One facet of this is errors in the data, and so we believe that using AED methods to analyse InstT datasets can potentially have a positive impact on LLM users.
However, the demographics of all annotators are fairly uniform, and yet in some cases there was substantial disagreement on what constitutes an error.
Therefore, we believe that a broader discussion involving more stakeholders is needed to get a diverse perspective on what is the desired behaviour of LLMs and thus what constitutes an error in InstT datasets.

\section*{Acknowledgments}
We thank Shijia Zhou for helping with the annotation of SNI-Donkii. Many thanks to the members of MaiNLP for their comments earlier drafts of on the paper. This research is in parts supported by European Research Council (ERC) grant agreement No.\ 101043235.

\bibliography{anthology,custom}

\appendix

\section{Donkii Data Statement} \label{sec:data_statement}
Following~\cite{bender-friedman-2018-data}, the following outlines the data statement for Donkii:
\begin{itemize}
\item \textsc{A. CURATION RATIONALE} Enrichment of existing instruction-tuning datasets with annotations for erroneous instances
\item \textsc{B. LANGUAGE VARIETY} English with the exact variant(s) unkown because of the large number of different sources of data
\item \textsc{C. SPEAKER DEMOGRAPHIC} Unknown because of the large number of different data sources
\item \textsc{D. ANNOTATOR DEMOGRAPHIC} Three post-doctoral researchers and one Master's student (age: 25-40),  gender: male and female. Native language: Russian, German. Socioeconomic status: higher-education student and university researchers.
\item \textsc{E. SPEECH SITUATION} Unknown because of the large number of different data sources
\item \textsc{F. TEXT CHARACTERISTICS}  Unknown because of the large number of different data sources
\item \textsc{PROVENANCE APPENDIX}
\begin{itemize}
    \item Alpaca~\citep{taori-etal-2023-alpaca}, CC By NC 4.0, \url{https://github.com/tatsu-lab/stanford_alpaca}
    \item AlpacaDataCleaned~\citep{ruebsamen-etal-2023-adc}, Apache 2.0, \url{https://github.com/gururise/AlpacaDataCleaned}
    \item Public Pool of Prompts~\citep{sanhMultitaskPromptedTraining2022}, Apache 2.0, \url{https://huggingface.co/datasets/bigscience/P3}
    \item Super-Natural Instructions~\citep{wang-etal-2022-super}, Apache 2.0, \url{https://github.com/allenai/natural-instructions}
\end{itemize}

\end{itemize}

\section{SNI Annotation Guidelines} \label{sec:guidelines_sni}
For this annotation effort, we assume a pair-wise annotation setting. You are shown two tasks and the instances that differ between them. You should judge whether one of the two tasks contains fewer errors than the other one. Each task has the following fields:
\begin{itemize}
\item \textbf{Definition}:  The instruction to the language model. E.g. ‘solve the following equation for x’
\item \textbf{Instances}: Each with the following fields:
\item \textbf{Input}: The input complementing the instruction. E.g. ‘equation: x + 2 = 5’. 
Output: The gold-standard output expected from the model
\end{itemize}

There are four possible labels:
\begin{enumerate}
\item A is better than B
\item B is better than A
\item A and B are the same
\item I don’t know
\end{enumerate}

Additionally, there is a field for short free-form comments where you can (but don’t have to) note a reason for your annotation.

We assume that the dataset is used to train a current text-only vanilla LLM like GPT3 or Llama. That is, it does not have access to tools and cannot process multi-modal input.

The following rules apply for differences between tasks A and B. We assume that the difference mentioned in the rule is the only difference between both (ceteris paribus). If more than one rule applies, we leave the choice to the best judgment of the annotator. The goal is to make only relative judgments for the given pair without considering the “absolute quality” of the instances. Even when both contain very little or many errors, if B clearly contains more significant errors than A, this should be annotated as “A is better than B”.

Rules:
\begin{itemize}
\item If B contains more errors than A, but those are only few and thus don’t affect the majority of the instances differing between A and B, then A and B are equal. As a guideline: If more not more than 90\% contain the error, then they probably should be equal.
\item Be lenient in your annotations. If you are unsure whether something is an error, then better go for A and B are equal.

\item \textbf{Factual correctness}: If the output of A can be interpreted as factually correct, but the one in B cannot, then A is better than B.
Example: 
\begin{itemize}
    \item Instruction: Tell me the title of the most popular song released in 2020 so far.
    \item Output A: The most popular song released in 2020 so far is "Blinding Lights" by The Weeknd.
    \item Output B:  The most popular song released in 2020 so far is "The Box" by Roddy Rich.
    \item Explanation: A is better than B, because, while the answer to A is ambiguous (there are multiple measures of popularity), “The Box” was released in 2019 and thus is clearly wrong.
\end{itemize}

\item \textbf{Noise}: If B contains noise (e.g. technical artifacts) but A does not, then A is better than B.
Example:
\begin{itemize}
\item Instruction: Suggest the best strategy for a five-second TV commercial.
\item Input A: 
\item Input B: “NoInput”
\item Explanation: A is better than B, because “NoInput” is clearly a technical artifact (even despite A being empty - i.e. no output better than noise).
\end{itemize}

\item \textbf{Only output}: Judge A and B based on the output field not instruction or input. 
Justification: It is not clear whether low-quality input with high-quality output improves or diminishes instruction tuning performance.
Example:
\begin{itemize}
\item Instruction A: Convert the following number in hexadecimal format.
\item Input A: 18
\item Instruction B: Convert the number 18 to hexadecimal
\item Input B:
\item Explanation B: A and B are equal, even though one could prefer A over B because input and instruction are cleanly separated.
\end{itemize}

\item \textbf{Unclear instruction}: If it is impossible to guess user intent based on the instruction in B, but it is possible to guess it in A, then A is better than B. 
Example:
\begin{itemize}
\item Instruction A: Find the average value of the following list of numbers
\item Instruction B: Process the following data and output the results
\item Input: List: [3, 7, 2, 5]
\item Explanation: A is better than B because for B it is not clear at all how the model should process the data. 
\end{itemize}

\item \textbf{Tool usage}:
If B requires tool usage (e.g. access to a search engine) but A doesn’t, then A is better than B. Justification: We assume that the dataset is used to instruction-tune a vanilla LM without access to tools.
Example:
\begin{itemize}
\item Instruction A: Provide a brief overview about the following topic.
\item Input A: Volcanology
\item Instruction B: Take a Wikipedia article and rewrite it in your own words.
\item Input B: https://en.wikipedia.org/wiki/Volcanology
\item Explanation: A is better than B because B requires access to a web browser.
\end{itemize}

\item \textbf{Multi-modal input}:
If B contains multi-modal input (e.g. an image file) but A doesn’t, then A is better than B.  Justification: We assume that the dataset is used to instruction-tune a vanilla text-only LM.
Example:
\begin{itemize}
\item Instruction: Critique the given painting.
\item Input A: The painting is an abstract composition of vibrant yellow, blue, and pink hues that appear in an haphazard, yet balanced form and serve as an evocation of life, joy, and emotion.
\item Input B: [Painting attached]
\item Explanation: A is better than B because B contains multimodal input.
\end{itemize}

\item \textbf{Temporal knowledge}: If B contains temporal knowledge but A doesn’t, then A is better than B. Justification: We want the instruction-tuned model to handle temporal knowledge gracefully.
Example:
\begin{itemize}
\item Instruction A: What is the name of the 46th president of the United States?
\item Instruction B: What is the name of the current president of the United States?
\item Explanation: A is better than B because the answer to B will change over time while the answer to A is static.
\end{itemize}

\item \textbf{Formatting}: If A and B differ only in formatting, then A and B are equal
Example:
\begin{itemize}
\item Output A: 
- Astonished
- Amazed
- Shocked
- Stunned
- Speechless
- Bewildered"
\item Output B: Astonished, amazed, shocked, stunned, speechless, bewildered.
Explanation: A is equal to B because the output only differs in formatting
\end{itemize}

\end{itemize}

\section{ADC Annotation Guidelines} \label{sec:guidelines_adc}

For this annotation effort, we assume a pair-wise annotation setting. You are shown two instances and have to judge which of both would you preferably include in an instruction-tuning dataset. Each instance has two to three fields: 
\begin{itemize}
\item \textbf{Instruction}: The instruction to the language model. E.g. ‘solve the following equation for x’
\item \textbf{Input} (optional): The input complementing the instruction. E.g. ‘equation: x + 2 = 5’. Instructions can be self-contained, thus Input is optional.
\item \textbf{Output}: The gold-standard output expected from the model
\end{itemize}

There are four possible labels:
\begin{enumerate}
    \item  A is better than B
    \item B is better than A
    \item A and B are the same
    \item  I don’t know
\end{enumerate}

Additionally, there is a field for short free-form comments where you can (but don’t have to) note a reason for your annotation.

We assume that the dataset is used to train a current text-only vanilla LLM like GPT3 or Llama. That is, it does not have access to tools and cannot process multi-modal input.

The following rules apply for differences between instances A and B. We assume that the difference mentioned in the rule is the only difference between both (ceteris paribus). If more than one rule applies, we leave the choice to the best judgment of the annotator. The goal is to make only relative judgments for the given pair without considering the “absolute quality” of the instances. Even when both are very high or low quality, if B is clearly worse than A, this should be annotated as “A is better than B”.

Rules:

\begin{itemize}

\item \textbf{Factual correctness}: If the output of A can be interpreted as factually correct, but the one in B cannot, then A is better than B.
Example: 
\begin{itemize}
    \item Instruction: Tell me the title of the most popular song released in 2020 so far.
    \item Output A: The most popular song released in 2020 so far is "Blinding Lights" by The Weeknd.
    \item Output B:  The most popular song released in 2020 so far is "The Box" by Roddy Rich.
    \item Explanation: A is better than B, because, while the answer to A is ambiguous (there are multiple measures of popularity), “The Box” was released in 2019 and thus is clearly wrong.
\end{itemize}

\item \textbf{Noise}: If B contains noise (e.g. technical artifacts) but A does not, then A is better than B.
Example:
\begin{itemize}
\item Instruction: Suggest the best strategy for a five-second TV commercial.
\item Input A: 
\item Input B: “NoInput”
\item Explanation: A is better than B, because “NoInput” is clearly a technical artifact (even despite A being empty - i.e. no output better than noise).
\end{itemize}

\item \textbf{Only output}: Judge A and B based on the output field not instruction or input. 
Justification: It is not clear whether low-quality input with high-quality output improves or diminishes instruction tuning performance.
Example:
\begin{itemize}
\item Instruction A: Convert the following number in hexadecimal format.
\item Input A: 18
\item Instruction B: Convert the number 18 to hexadecimal
\item Input B:
\item Explanation B: A and B are equal, even though one could prefer A over B because input and instruction are cleanly separated.
\end{itemize}

\item \textbf{Unclear instruction}: If it is impossible to guess user intent based on the instruction in B, but it is possible to guess it in A, then A is better than B. 
Example:
\begin{itemize}
\item Instruction A: Find the average value of the following list of numbers
\item Instruction B: Process the following data and output the results
\item Input: List: [3, 7, 2, 5]
\item Explanation: A is better than B because for B it is not clear at all how the model should process the data. 
\end{itemize}

\item \textbf{Tool usage}:
If B requires tool usage (e.g. access to a search engine) but A doesn’t, then A is better than B. Justification: We assume that the dataset is used to instruction-tune a vanilla LM without access to tools.
Example:
\begin{itemize}
\item Instruction A: Provide a brief overview about the following topic.
\item Input A: Volcanology
\item Instruction B: Take a Wikipedia article and rewrite it in your own words.
\item Input B: https://en.wikipedia.org/wiki/Volcanology
\item Explanation: A is better than B because B requires access to a web browser.
\end{itemize}

\item \textbf{Multi-modal input}:
If B contains multi-modal input (e.g. an image file) but A doesn’t, then A is better than B.  Justification: We assume that the dataset is used to instruction-tune a vanilla text-only LM.
Example:
\begin{itemize}
\item Instruction: Critique the given painting.
\item Input A: The painting is an abstract composition of vibrant yellow, blue, and pink hues that appear in an haphazard, yet balanced form and serve as an evocation of life, joy, and emotion.
\item Input B: [Painting attached]
\item Explanation: A is better than B because B contains multimodal input.
\end{itemize}

\item \textbf{Temporal knowledge}: If B contains temporal knowledge but A doesn’t, then A is better than B. Justification: We want the instruction-tuned model to handle temporal knowledge gracefully.
Example:
\begin{itemize}
\item Instruction A: What is the name of the 46th president of the United States?
\item Instruction B: What is the name of the current president of the United States?
\item Explanation: A is better than B because the answer to B will change over time while the answer to A is static.
\end{itemize}

\item \textbf{Formatting}: If A and B differ only in formatting, then A and B are equal
Example:
\begin{itemize}
\item Output A: 
- Astonished
- Amazed
- Shocked
- Stunned
- Speechless
- Bewildered"
\item Output B: Astonished, amazed, shocked, stunned, speechless, bewildered.
Explanation: A is equal to B because the output only differs in formatting
\end{itemize}

\item \textbf{Global properties}: If A and B differ only with respect to the full dataset, e.g. because A increases grammatical diversity in the input but B doesn’t, then A and B are equal
Example:
\begin{itemize}
\item Instruction A: The average of 10 numbers is 85. If the numbers 70 and 76 are removed from the set of numbers, what is the average of the remaining numbers?
\item Instruction B: Find the average of 85, 44 and 102.
\item Explanation: A and B are equal, even though more complex problems as A are much less frequent in the whole dataset than problems of the type B.

\end{itemize}

\item \textbf{Subjectivity}: If the annotator feels that their preference for one instance is strongly subjective, then A and B are equal.

\item \textbf{Leniency}: Be lenient in your annotations. If you are unsure whether something is an error, then better go for A and B are equal. If both outputs can be interpreted as correct, then A and B are equal

\end{itemize}

\clearpage

\onecolumn

\begin{table*}[htpb]
\section{Examples of errors} \label{sec:error_appendix}
\adjustbox{{max width=0.9\linewidth}}{%
\centering
\small
\begin{tabular}{clp{5cm}p{5cm}p{5cm}}
\toprule
Error & Source & Instruction & Input & Output (shortened) \\
\midrule

\multicolumn{5}{c}{\textcolor{output}{\textbf{Incorrect output}}} \\ \midrule

\makecell{ \textcolor{output}{Wrong} \\ \textcolor{output}{output }}&   SNI  &  Given a sentence and an entity, the task is to select the authors sentiment towards the enity. Sentiments can be Positive, Neutral and Negative. <...> &  What is the sentiment of the following document towards the entity Hayley Smith ? Hayley Smith was diagnosed with chronic depression in  her  early twenties: ``I'd been bottling up quite a bit through most of my teens. Then it hit me a like a brick wall ''  she  said. & \textcolor{output}{ Neutral } \\
\makecell{ \textcolor{output}{Empty} \\ \textcolor{output}{output}} &  Alpaca   &  Create a flow chart to explain the working of a web browser.  &   &  \\
\makecell{ \textcolor{output}{Labels} \\ \textcolor{output}{flipped}} &   SNI & In this task you will be given a passage and a yes/no question based on the passage. You should answer the question using the information from the passage.   &  Superfecundation is the fertilization of two or more ova from the same cycle by sperm from separate acts of sexual intercourse, which can lead to twin babies from two separate biological fathers. [...]  question: can a woman produce twins of different fathers?& 	\textcolor{output}{No} \\
\makecell{ \textcolor{output}{Output} \\  \textcolor{output}{space}} &    SNI & You are given a science question (easy-level) and four answer options (associated with "A", "B", "C", "D"). Your task is to find the correct answer based on scientific facts, knowledge and reasoning. & Which group of organisms can show significant trait changes in the shortest period of time?  
 (A) bacteria (B) birds (C) fish (D) reptiles  & \textcolor{output}{1}   \\
\makecell{ \textcolor{output}{Off-} \\ \textcolor{output}{instruction}} & Alpaca     & Given three numbers (x, y, z), find the greatest common factor.   &  2, 4, 2008 & 		The greatest common factor of \textcolor{output}{2, 4, and 8} is 2. \\
\makecell{\textcolor{output}{Grammar} \\ \textcolor{output}{error}} &  SNI  & In this task, you are given a question and an answer, you would be asked to create the sentence based on the Question-Answer provided. It should be contained within the Question-Answer provided.  & Question: What is the most common type of cancer in adult males? Answer: prostate	  &  \textcolor{output}{Prostate} is the most common type of cancer in adult males.    \\

\midrule
\multicolumn{5}{c}{\textcolor{factual}{\textbf{Factual Errors}}} \\ 
\midrule

\makecell{\textcolor{factual}{Temporal} \\ \textcolor{factual}{context}} &   Alpaca  &  What is the average \textcolor{factual}{annual} salary for a customer service developer?  & &		The average \textcolor{factual}{annual} salary for a customer service developer is usually around \$74,508.     \\
\textcolor{factual}{Bad math} & Alpaca & Calculate the average of the following three numbers: 10, 12, 15. &   &  The average of 10, 12, and 15 is \textcolor{factual}{12.67}.  \\
\textcolor{factual}{Encyclopedic} & Alpaca &  What is the average air pressure of a human's lungs?	&  & 	\textcolor{factual}{The average air pressure in a human's lungs is between four and seven centimeters of water (4–7 cm H2O).}    \\

\midrule
\multicolumn{5}{c}{\textcolor{noise}{\textbf{Noise}}} \\ 
\midrule
\makecell{\textcolor{noise}{Noise}} &  Alpaca   & Write an essay about the benefits of public transportation.	  &   \textcolor{noise}{No input required.}  &  Public transportation is an invaluable resource for communities and cities around the world. <...> \\

\midrule
\multicolumn{5}{c}{\textcolor{input}{\textbf{Underspecified Input
}}} \\ 
\midrule
\makecell{\textcolor{input}{Missing} \\ \textcolor{input}{context}  \\ \textcolor{input}{information}} &  Alpaca   & Write a tweet summarizing this article about an animal rescue in Thailand. &	\textcolor{input}{The article}	& Amazing news! An animal rescue in Thailand has successfully freed 63 bears from animal trafficking. The rescued bears will now receive the proper care and attention they need.   \\
\makecell{\textcolor{input}{Unclear} \\ \textcolor{input}{instruction}} &  Alpaca   &  \textcolor{input}{Sort the following 10 fruits into three categories.}  & Bananas, Apples, Pineapples, Plums, Cherries, Peaches, Grapes, Watermelons, Mangoes, Oranges  & Category 1: Bananas, Apples, Pineapples, Plums, Cherries;
Category 2: Peaches, Grapes, Watermelons; Category 3: Mangoes, Oranges \\

\midrule
\multicolumn{5}{c}{\textcolor{modality}{\textbf{Model-modality mismatch}}} \\ 
\midrule
\makecell{\textcolor{modality}{Requires} \\ \textcolor{modality}{web} \\ \textcolor{modality}{browsing}} & Alpaca    &  Provide a summarization of the article.	  &  \textcolor{modality}{\texttt{https://en.wikipedia.org/wiki/ Artificial\_intelligence}} &  Artificial Intelligence, commonly known as AI, refers to the simulation of human intelligence processes by machines, especially computer systems. <...>  \\
\makecell{\textcolor{modality}{Table} \\  \textcolor{modality}{in input}} &  Alpaca   &   &  \textcolor{modality}{Dataset of 500 people with the variable ``income''.}  &  The mean income is \$50,000. \\
\makecell{\textcolor{modality}{Image} \\  \textcolor{modality}{in input}} & Alpaca &  How would you use this photo in a marketing campaign?   &   \textcolor{modality}{<image included in email>}  &  The photo can be used to create a compelling marketing campaign that draws attention to the product or service. <...>  \\
\makecell{\textcolor{modality}{Image} \\  \textcolor{modality}{in output}}&  Alpaca  &  Make a word cloud on the given topic.	 &  Artificial Intelligence	  & \textcolor{modality}{<Word Cloud Output>} \\

\midrule
\multicolumn{5}{c}{\textcolor{formatting}{\textbf{Formatting}}} \\ 
\midrule
\makecell{\textcolor{formatting}{Formatting}} &  SNI   &  In this task, you will be given a short story. One sentence from the story is chosen. Consider the events that happen before that sentence, or are likely to have happened before it. Does any of them directly cause it, or simply make it possible? You should write your answer in the form " A >causes/enables> B". Try to use phrases and sentences from the story to compose your answer when possible. & story: I went down to the tidepool to watch the tide roll out. I sat on the dock and waited, while listening to my mp3 player. Once the tide was out, I saw Something shiny in the muddy bottoms. I went down and found that it was a gold ring! Today was my lucky day!
 selected sentence: I went down to the tidepool to watch the tide roll out. &  \textcolor{formatting}{ I decide togotothe tidepool >Causes/Enables> I gotothe tidepool } \\

\bottomrule
\end{tabular}%
}
\caption{Examples of errors.}
\label{tab:more_examples}%
\end{table*}%

\clearpage

\twocolumn
\section{Hyperparameters} \label{sec:hyperparameters}
We experiment with four sizes, namely small (60 million parameters), base (220 million), large (770 million), and 3B (3 billion) using NVIDIA A100 cards.
We train each of the models for 10 epochs as a seq2seq LM using a batch size of 60 and a learning rate of $1e-3$.
Note, that we train separate models for each of the three datasets and leave the exploration of possible synergies across datasets for future work.
We set the maximum source length to 512/768/768 for P3-Donkii/SNI-Donkii/APC-Donkii and the output length to 256.

\clearpage

\section{Preliminary experiments on the impact of errors on downstream performance}
\label{sec:impact}
We conduct a preliminary experiment on how errors in InstT datasets affect downstream performance in a case study.
For this, we use the training and evaluation setup of Tk-Instruct~\citep{wang-etal-2022-super}, which is the main model trained on SNI using the code provided by the authors.\footnote{\url{https://github.com/yizhongw/Tk-Instruct}}
To investigate the effect of errors, we contrast two models: $\text{Tk-Instruct}_\text{err}$ and $\text{Tk-Instruct}^*$.
Both models are based on the three billion parameter version of T5. 
For $\text{Tk-Instruct}_\text{err}$, we use all 17 tasks that we found to be erroneous in SNI-Donkii.
To these, we add a sample of an additional 100 tasks from the original Tk-Instruct training data.
This results in a training data set of $6,985$ instances across 117 tasks and an error rate of approximately 8\%.
For $\text{Tk-Instruct}^*$ we replace all erroneous instances with corrected instances from the same task.
We adapt the training pipeline to our limited computational budget: 
We train and evaluate the model in a strict zero-shot setting without providing few-shot examples to reduce the input length of the instances.
Second, we use only at most 64 instances per task, because \citet{wang-etal-2022-super} find that increasing this number does not improve performance.
We train the model for 30 epochs with a batch size of 1024.
We use the same held-out task mixture for evaluation as \citet{wang-etal-2022-super} but remove all tasks that are in our training data.
To evaluate the impact of errors on instruction tuning, we follow \citet{wang-etal-2022-super} and use RougeL for evaluation.
Surprisingly, we find that the difference between the two models is small.
$\text{Tk-Instruct}^*$ achieves an overall RougeL score of $35.9\%$, while $\text{Tk-Instruct}_\text{err}$ achieves $35.7\%$.
Moreover, $\text{Tk-Instruct}_\text{err}$ even generates the correct answer for instances where it observed incorrect answers during training.
Both observations suggest that instruction-tuned models may be robust to small numbers of errors in their training data.
However, when we prompt the published version of T0\footnote{\url{https://huggingface.co/bigscience/T0_3B}}, the model trained on P3, with a prompt template for which during training it erroneously always observed empty strings as output\footnote{"Question 1: [...]? Question 2: [...]? Do these questions convey the same meaning? Yes or no?"}, we find that it will always respond with an empty string.
This motivates further research into when and how errors in InstT datasets propagate into models.

\end{document}